\documentclass[letterpaper]{article} 
\usepackage{aaai24}  
\usepackage{times}  
\usepackage{helvet}  
\usepackage{courier}  
\usepackage[hyphens]{url}  
\usepackage{graphicx}
\usepackage{natbib}  
\usepackage{caption} 
\usepackage{algorithm}
\usepackage{algorithmic}

\usepackage{newfloat}
\usepackage{listings}
\DeclareCaptionStyle{ruled}{labelfont=normalfont,labelsep=colon,strut=off} 
\floatstyle{ruled}
\newfloat{listing}{tb}{lst}{}
\floatname{listing}{Listing}
\usepackage{amsmath}
\usepackage{amssymb}
\usepackage{mathtools}
\usepackage{amsthm}

\usepackage[capitalize,noabbrev]{cleveref}

\theoremstyle{plain}

\theoremstyle{definition}

\theoremstyle{remark}

\usepackage[textsize=tiny]{todonotes}

\usepackage{amsmath,amsfonts,bm}

\def\eqref#1{equation~\ref{#1}}

\def\1{\bm{1}}

\DeclareMathAlphabet{\mathsfit}{\encodingdefault}{\sfdefault}{m}{sl}
\SetMathAlphabet{\mathsfit}{bold}{\encodingdefault}{\sfdefault}{bx}{n}

\newcommand{\R}{\mathbb{R}}

\usepackage{refcount}
\usepackage{url}
\usepackage{float}
\usepackage{xcolor,colortbl}
\definecolor{Gray}{gray}{0.85}
\definecolor{LightCyan}{rgb}{0.85,0.92,0.90}

\usepackage{subfig}
\usepackage{xspace}
\usepackage{graphicx}
\usepackage{caption}

\usepackage{multirow}
\usepackage{enumitem}
\usepackage{multicol}

\usepackage{booktabs}

\title{ Diversified Ensemble of Independent Sub-Networks for\\ Robust Self-Supervised Representation Learning}

\author {
    Amirhossein Vahidi\textsuperscript{\rm 1,2},
    Lisa Wimmer\textsuperscript{\rm 1,2},
    Hüseyin Anil Gündüz\textsuperscript{\rm 1,2},
    Bernd Bischl\textsuperscript{\rm 1,2},
    Eyke Hüllermeier\textsuperscript{\rm 3},
    Mina Rezaei\textsuperscript{\rm 1,2}
}
\affiliations {
    \textsuperscript{\rm 1}Institute of Statistics, LMU Munich \quad \textsuperscript{\rm 2}Munich Center for Machine Learning \quad \textsuperscript{\rm 3}Institute of Informatics, LMU Munich 
}

\usepackage{bibentry}

\begin{document}

\maketitle

\begin{abstract}

Ensembling a neural network is a widely recognized approach to enhance model performance, estimate uncertainty, and improve robustness in deep supervised learning. However, deep ensembles often come with high computational costs and memory demands. In addition, the efficiency of a deep ensemble is related to diversity among the ensemble members which is challenging for large, over-parameterized deep neural networks. Moreover, ensemble learning has not yet seen such widespread adoption, and it remains a challenging endeavor for self-supervised or unsupervised representation learning.
Motivated by these challenges, we present a novel self-supervised training regime that leverages an ensemble of independent sub-networks, complemented by a new loss function designed to encourage diversity. Our method efficiently builds a sub-model ensemble with high diversity, leading to well-calibrated estimates of model uncertainty, all achieved with minimal computational overhead compared to traditional deep self-supervised ensembles.
To evaluate the effectiveness of our approach, we conducted extensive experiments across various tasks, including in-distribution generalization, out-of-distribution detection, dataset corruption, and semi-supervised settings. The results demonstrate that our method significantly improves prediction reliability. Our approach not only achieves excellent accuracy but also enhances calibration, surpassing baseline performance across a wide range of self-supervised architectures in computer vision, natural language processing, and genomics data.

\end{abstract}

\section{Introduction}

Ensemble learning has become a potent strategy for en\-han\-cing model performance in deep learning \cite{hansen1990neural,dietterich2000ensemble,Lakshminarayanan2017}. This method involves combining the outputs of multiple independently-trained neural networks, all using the same architecture and same training dataset but differing in the randomness of their initialization and/or training. Despite its remarkable effectiveness, training deep ensemble models poses several challenges: i) The high performance achieved by deep ensembles comes with a significant increase in computational costs. Running multiple neural networks independently demands more resources and time. ii) Maintaining diversity among ensemble members -- a property often critical to success -- becomes progressively difficult for large, over-parameterized deep neural networks~\cite{dice2021,dabouei2020exploiting} in which the main source of diversity comes from random weight initialization. iii) Most of the existing literature focuses on deep ensembles for supervised models. Adapting these approaches to unsupervised and self-supervised models requires careful consideration and evaluation to ensure comparable performance.

 In recent years, self-supervised learning methods have achieved cutting-edge performance across a wide range of tasks in natural language processing (NLP; ~\citep{devlin2018bert,brown2020language}, computer vision~\citep{chen2020simple, bardes2021vicreg,grill2020bootstrap,rezaei2021deep,lienen2022conformal}, multimodal learning~\citep{radford2021learning, li2022clip,shi2022learning}, and bioinformatics~\citep{gunduz2021self}. In contrast to supervised techniques, these models learn representations of the data without relying on costly human annotation. Despite remarkable progress in recent years, self-supervised models do not allow practitioners to inspect the model's confidence. This problem is non-trivial given the degree to which critical applications rely on self-supervised methods. As recently discussed by LeCun\footnote{https://ai.facebook.com/blog/self-supervised-learning-the-dark-matter-of-intelligence/}, representing predictive uncertainty is particularly difficult in self-supervised contrastive learning for computer vision. Therefore, quantifying the predictive uncertainty of self-supervised models is critical to more reliable downstream tasks. Here, we follow the definition of reliability as described by Plex~\citep{tran2022plex}, in which the ability of a model to work consistently across many tasks is assessed. In particular, \citet{tran2022plex} introduce three general desiderata of reliable machine learning systems: a model should generalize robustly to \emph{new tasks}, as well as \emph{new datasets}, and represent the associated \emph{uncertainty} in a faithful manner.

In this paper, we introduce a novel, robust, and scalable framework for ensembling \emph{self-supervised learning} while \emph{preserving performance} with a negligible increase in computational cost and \emph{encouraging diversity among the ensemble of sub-networks}.


Our contributions can be summarized as follows:
\begin{itemize}
    \item We propose a novel scalable ensemble of self-supervised learning to be robust, efficient, and enhance the model performance in various downstream tasks.
    \item We develop a complementary loss function to enforce diversity among the independent sub-networks. 
    \item We perform extensive empirical analyses to highlight the benefits of our approach. We demonstrate that this inexpensive modification achieves very competitive (in most cases, better) predictive performance: 1) on in-distribution (IND) and out-of-distribution (OOD) tasks; 2) in semi-supervised settings; 3) learns a better predictive performance-uncertainty trade-off than compared baselines (i.e., exhibits high predictive performance and low uncertainty on IND datasets as well as high predictive performance and high uncertainty on OOD datasets). 
    \vspace{-3pt}
\end{itemize}

\begin{figure*}
\centering
    \includegraphics[width=0.66\textwidth]{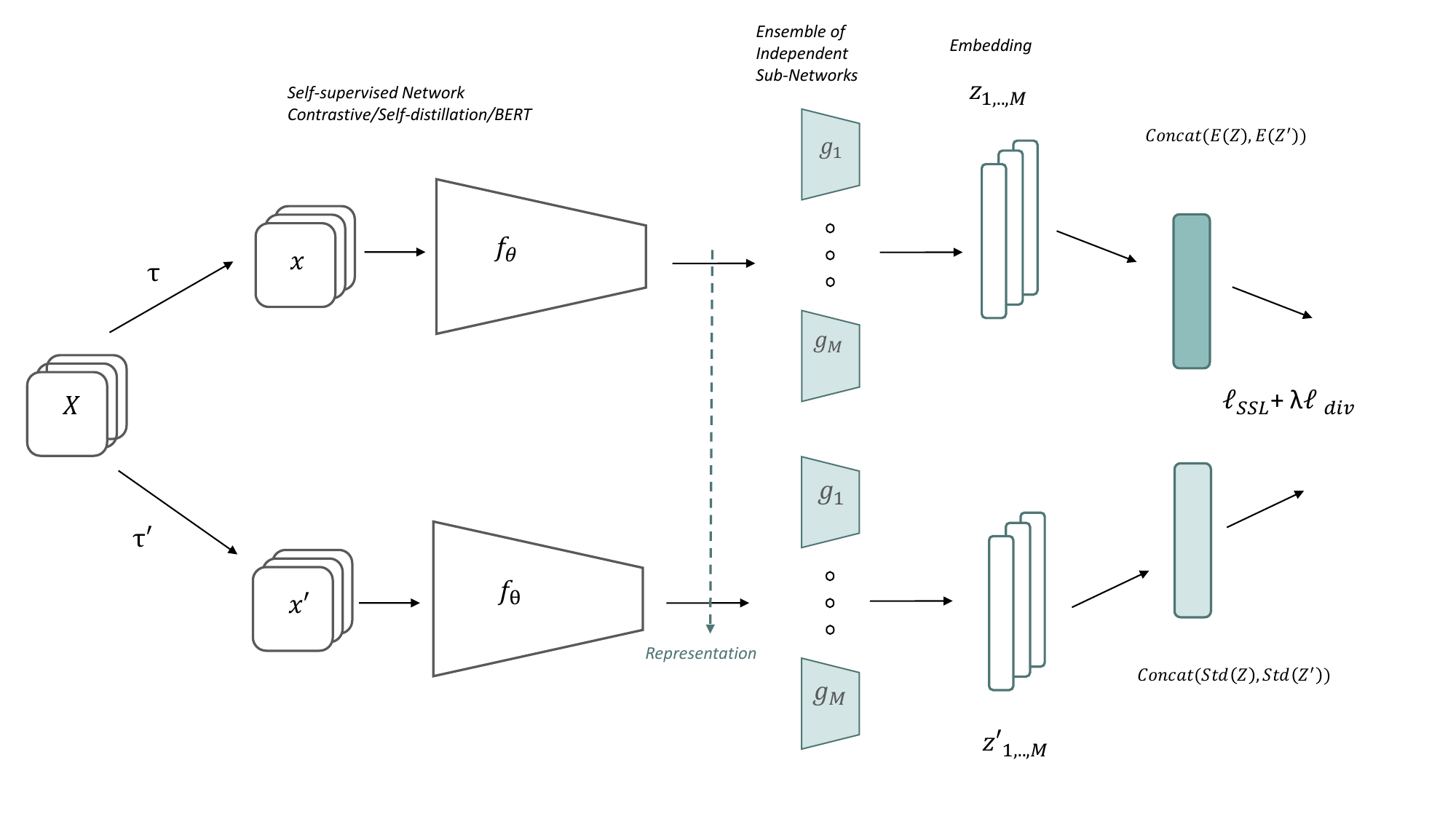}
    \vspace{-12pt}
    \caption{ Illustration of our proposed method. Given a batch $\bm{X}$ of input samples, two different views $\tilde{\bm{x}}$ and $\tilde{\bm{x}}^\prime$ are produced for each sample, which is then encoded into representations by the encoder network $f_{\bm{\theta}^\prime}$. The representations are projected to the ensemble of independent sub-networks $g_m$, where each sub-network produces embedding vectors $\bm{z}$ and $\bm{z}^\prime$. The mean value of these embeddings is passed to the self-supervised loss, while their standard deviation is used for the diversity loss. Finally, the total loss is computed by a combination of the two loss components.} 
\label{fig:method}
    \vspace{-9pt}
\end{figure*}

\section{Related Work} 

\noindent\textbf{Self-supervised learning ~} 
For most large-scale modeling problems, learning under full supervision is severely inhibited by the scarcity of annotated samples. Self-supervised learning techniques, which solve \emph{pretext tasks} \citep{devlin2018bert} to generate labels from (typically abundant) unlabeled data, have proven to be a powerful remedy to this bottleneck. The learned feature maps can serve as a starting point for \emph{downstream} supervised tasks, such as classification, object detection, or sentiment analysis, with a substantially reduced need for labeled examples \citep{jaiswal_2020_SurveyContrastiveSelfSupervised}. Alternatively, the downstream application may directly use the extracted representation for problems such as anomaly OOD detection. While there have been attempts to make pretraining more robust by preventing embedding collapse \citep{bardes2021vicreg,rezaei2021deep} or boosting performance in OOD detection \citep{winkens_2020_ContrastiveTrainingImproved, sehwag_2021_SSDUnifiedFramework,rezaei2022bayesian,tran2022plex}, the aspect of \emph{uncertainty-awareness} has been studied to a lesser extent in the self-supervised context. Motivated by this, we present a simple way to make self-supervised learning robust during pretext-task learning.

\noindent\textbf{Ensemble learning ~}
Deep Ensembles~\citep{Lakshminarayanan2017} comprise a set of $M$ neural networks that independently train on the same data using random initialization. Deep ensembles often outperform other approaches in terms of calibration and predictive accuracy \citep{Ovadia2019,Gustafsson2020,rezaei2022deep,Ashukha2020,hullermeier2021aleatoric}, but their naive application incurs high computational complexity, as training, memory, and inference cost multiplies with the number of base learners. BatchEnsemble \citep{wen2020batchensemble} introduces multiple low-rank matrices with little training and storage demand, whose Hadamard products with a shared global weight matrix mimic an ensemble of models. Masksensemble \citep{durasov2021masksembles} builds upon Monte Carlo dropout \citep{gal2016dropout} and proposes a learnable (rather than a random) selection of masks used to drop specific network neurons.
MIMO \citep{mimo2021} uses ensembles of sub-networks diverging only at the beginning and end of the parent architecture -- thus sharing the vast majority of weights -- in order to obtain multiple predictions with a single forward pass. At test time, several copies of each sample are fed to the enlarged input layer, and the multi-head last layer returns an according number of predictions.
Although these methods reduce the inference time and computational resources required at training, the benefits are limited to the larger pretraining model that is used in self-supervised learning.

\noindent\textbf{Diversity in ensembles}: Diversity is a crucial component for successful ensembles. \citet{dice2021} classify existing approaches for encouraging diversity among ensemble members into three groups: i) methods that force \emph{diversity in gradients} with adaptive diversity in prediction~\citep{pang2019improving}, or using joint gradient phase and magnitude regularization (GPMR) between ensemble members \citep{dabouei2020exploiting}, ii) methods focusing on \emph{diversity in logits}, improving diversity with regularization and estimating the uncertainty of out-of-domain samples~\cite{liang2017enhancing}, or by bounding the Lipschitz constant of networks and limiting the variety of predictions against slight input changes~\cite{cisse2017parseval,tsuzuku2018lipschitz}, iii) methods promoting \textit{diversity in features} that increase diversity with adversarial loss \cite{chen2020learning} for conditional redundancy~\cite{dice2021}, information bottleneck \cite{sinha2021dibs,fischer2020conditional}, or $f1$-divergences~\cite{chen2020learning}. Our method belongs to this last category, where our loss function encourages the diversity of feature maps. 

\section{Method} \label{sec:method} 

We propose a simple principle to 1) make self-supervised pretraining robust with an ensemble of diverse sub-networks, 2) improve predictive performance during pretraining of self-supervised deep learning, 3) while keeping an efficient training pipeline. 

As depicted in Figure~\ref{fig:method}, our proposed method can be readily applied to the most recent trends in self-supervised learning~\citep{caron2021emerging,grill2020bootstrap,chen2020simple,devlinBert,gunduz2021self,klein2022scd} and is based on a joint embedding architecture. In the following sections, we first describe our proposed ensemble model, followed by the diversity loss, and then a discussion on diversity, and computational cost.

\subsection{Robust Self-Supervised Learning via Independent Sub-Networks}

\paragraph{Setting.} Given a randomly sampled mini-batch of data $\bm {X} =\{ \bm{x}_k \}_{k=1}^N \subset \mathcal{X} \subseteq \R^p$, the transformer function derives two augmented views $\tilde{\bm{x}}=\tau(\bm{x}), \tilde{\bm{x}}^\prime=\tau^\prime(\bm{x})$ for each sample in $\bm{X}$. The augmented views are obtained by sampling $\tau, \tau^\prime$ from a distribution over suitable data augmentations, such as masking parts of sequences~\citep{baevski2022data2vec,devlinBert}, partially masking image patches~\citep{he2022masked}, or applying image augmentation techniques~\citep{chen2020simple}.

The two augmented views $\tilde{\bm{x}}$ and $\tilde{\bm{x}}^\prime$ are then fed to an encoder network $f_{\bm{\theta}}$ with trainable parameters $\bm{\theta} \subseteq \R^d$. The encoder (e.g., ResNet-50~\citep{he2016deep}, ViT~\citep{dosovitskiy2020image}) maps the distorted samples to a set of corresponding features. We call the output of the encoder the \emph{representation}. Afterward, the representation features are transformed by $M$ independent sub-networks $\{ g_{\bm{\phi_m}} \}_{m=1}^M$ with trainable parameters $\bm{\phi}_m$ to improve the feature learning of the encoder network. The ensemble constructs from the representation $M$ different $q$-dimensional \emph{embedding} vectors $\{\bm{z}_m\}_{m=1}^M$, $\{\bm{z}_m^\prime\}_{m=1}^M$, respectively, for $\tilde{\bm{x}}$ and $\tilde{\bm{x}}^\prime$. We modify the conventional self-supervised loss and replace the usual $\bm{z}_m$ by the mean value $\bar{ \bm{z}} = (\bm{z}_1 + \ldots + \bm{z}_M)/M$, and similarly $\bm{z}_m^\prime$ by $\bar{ \bm{z}}^\prime$.
Averaging over the embeddings generated by the $M$ sub-networks is likely to increase robustness, which in turn may help to improve predictive performance in downstream tasks

\paragraph{Self-supervised loss.}
In the case of contrastive learning~\citep{chen2020simple}, the self-supervised loss $\ell_{\text{ssl}}$ with temperature $t > 0$ and cosine similarity $\mathrm{sim}(\cdot, \cdot)$ is computed as:
\begin{equation}\label{eq:loss:ssl}
    \ell_{\text{ssl}} \left(\tilde{\bm{x}}_k,\tilde{\bm{x}}^\prime_k \right) = -\log ~ \frac{\exp(\mathrm{sim}( \bar{ \bm{z}}_k, \bar{ \bm{z}}^\prime_k)/ t)}{ \sum_{i=1}^{2N} \mathbb{I}_{[k \neq i]} \exp(\mathrm{sim}(\bm \bar{ \bm{z}}_k, \bar{ \bm{z}}_i)/ t)}.
\end{equation}

\paragraph{Diversity loss.} Since diversity is a key component of successful model ensembles~\cite{fort2019deep}, we design a new loss function for encouraging diversity during the training of the sub-networks. We define the diversity regularization term $\ell_{\text{div}}$ as a hinge loss over the difference of the standard deviation across the embedding vectors $\{\bm{z}_{k, m}\}_{m=1}^M$, $\{\bm{z}_{k, m}^\prime\}_{m=1}^M$ to a minimum diversity of $\alpha > 0$. The standard deviation is the square root of the element-wise variance $\{\sigma_{k, o}^2 \}_{o=1}^q$:  

\begin{equation}~\label{eq:loss:var}
{\sigma}_{k, o}^2 = \tfrac{1}{M-1}   \textstyle\sum_{m=1}^M (z_{k, m, o} 
 - \bar{z}_{k, o})^2 + \epsilon \, ,  \nonumber 
\end{equation}

\noindent where we add a small scalar $\epsilon > 0$ to prevent numerical instabilities. The diversity regularization function is then given by:  

\begin{align} \label{eq:loss:div}
    \ell_{\text{div}}\left(\tilde{\bm{x}}_k, \tilde{\bm{x}}_k^\prime \right) = \textstyle \sum_{o = 1}^q & \operatorname{max}\left(0, \alpha - {\sigma_{k, o}} \right) \\ 
    & \, + \operatorname{max}(0, \alpha - {\sigma_{k, o}^\prime} ) \, , \nonumber
\end{align}

\noindent where $\sigma$ and $\sigma^\prime$ indicate standard deviation for the input sample and augmented views, respectively. 

\vspace{-3pt}
\begin{figure}[H]
\centering
    \includegraphics[width=0.49\textwidth]{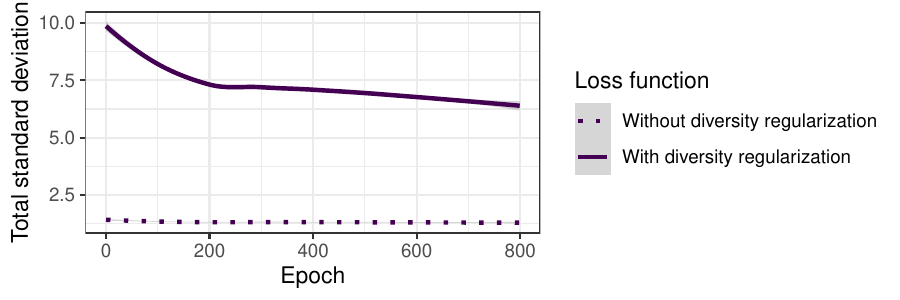}
    \vspace{-20pt}
  \caption{ \small \textbf{Total Standard Deviation}: sum of all standard deviations between independent sub-networks during training. Training with diversity loss (Eq.~\ref{eq:loss:div}) increases the standard deviation and improves the diversity between independent sub-networks. } 
\label{fig:loss}
\vspace{-7pt}
\end{figure}

\paragraph{Total loss.} The objective of the diversity loss is to encourage disagreement among sub-networks by enforcing the element-wise standard deviations to be close to $\alpha > 0$ and to thus prevent the embeddings from collapsing to the same vector. Figure~\ref{fig:loss} underlines the importance of the diversity loss on the total sum of standard deviations between different sub-networks, which increases by adding this loss. The total loss is calculated by combining the self-supervised loss (Eq.~\ref{eq:loss:ssl}) and the diversity loss (Eq.~\ref{eq:loss:div}), where the degree of regularization is controlled by a tunable hyperparameter $\lambda \geq 0$: 

\begin{equation} \label{eq:loss:total}
    \ell\left(\tilde{\bm{x}}_k, \tilde{\bm{x}}_k^\prime \right) = \ell_{\text{ssl}} \left(\tilde{\bm{x}}_k, \tilde{\bm{x}}_k^\prime \right)+ \lambda \cdot \ell_{\text{div}} \left(\tilde{\bm{x}}_k, \tilde{\bm{x}}_k^\prime \right).
\end{equation}
Finally, the total loss is aggregated over all the pairs in minibatch $\bm{X}$:

\begin{equation} \label{eq:simclr:loss}
    \mathcal{L}_{\text{total}}=
    \tfrac{1}{N} {\textstyle\sum_{k=1}^{N} \ell\left(\tilde{\bm{x}}_k, \tilde{\bm{x}}_k^\prime \right) }.
\end{equation}

\paragraph{Gradients.} Consider the output of the encoder $f_{\bm{\theta}}(\bm{x}) = b$ and the output of the $m$-th linear sub-network $\bm{z}_m = g_m(b) = w_m\cdot b$. The weight $w_m$ is updated by two components during backpropagation, the first of which depends on the self-supervised loss and is the same for the entire ensemble, while the second term depends on the diversity loss and is different for each sub-network. Given Eq.~\ref{eq:loss:var}, we simplify the equation by vector-wise multiplication since the sub-networks are linear; furthermore, we omit the numerical stability term since it does not have an effect on the derivative. The element-wise standard deviation can be computed as follows:

\begin{equation}\label{eq:simple:var}
     {\sigma}_{k,o} = \left(\tfrac{1}{M-1}  \textstyle\sum_{m=1}^M (\bm{z}_{k, m, o} - \bar{\bm{z}}_{k, o})^2\right)^{\tfrac{1}{2}}.
\end{equation}

Consider Eq.~\ref{eq:loss:div} for aggregating the element-wise standard deviations for one observation ($\bm{x}$) and assume ${\sigma}_{k} < \alpha$; otherwise, the diversity loss is zero when ${ \alpha \leq \sigma}_{k} $. The derivative of the loss with respect to $\bm{z}_{k,\hat{m}, o}$, $\hat{m} \in {1,\ldots,M} $, is then given as follows:

\begin{equation}\label{eq:derivative:loss}
{\frac{\partial \left( \ell_{\text{div}} \right) }{ \partial \bm{z}_{k,\hat{m}, o}}} = \frac{-A}{M-1} \cdot (\bm{z}_{k, \hat{m}, o} - \bar{\bm{z}}_{k, o}), 
\end{equation}
where $A := \tfrac{1}{M-1}  \textstyle\sum_{m=1}^M (\bm{z}_{k, m, o} - \bar{\bm{z}}_{k, o})^2)$. The proof is provided in the appendix (see Theoretical Supplement). 

In the optimization step of stochastic gradient descent (SGD), the weight of sub-network $\hat{m}$ is updated by:

\begin{equation}\label{eq:update:SGD}
\eta \cdot \nabla_{w_{\hat{m},o}} \ell_{\text{div}} = -C \cdot (\bm{z}_{k,\hat{m}, o} - \bar{\bm{z}}_{k, o}),
\end{equation}

where $\eta > 0$ is the learning rate, and $C$ is constant with respect to $w_{\hat{m}, o}$, which depends on the learning rate, number of sub-networks, $A$, and $b$.
The proof is provided in Appendix (see Theoretical Supplement). 

Eq.~\ref{eq:update:SGD} shows the updating step in backpropagation. Hyperparameter $\alpha$ prevents ${\bm{z}}_{k,\hat{m}, o}$ from collapsing to the a single point. Hence, $w_{\hat{m},o}$ is updated in the opposite direction of $\bar{\bm{z}}_{k,o}$, so the diversity loss prevents weights in the sub-networks from converging to the same values.

\begin{figure}[!t]
\includegraphics[width=0.43\textwidth]{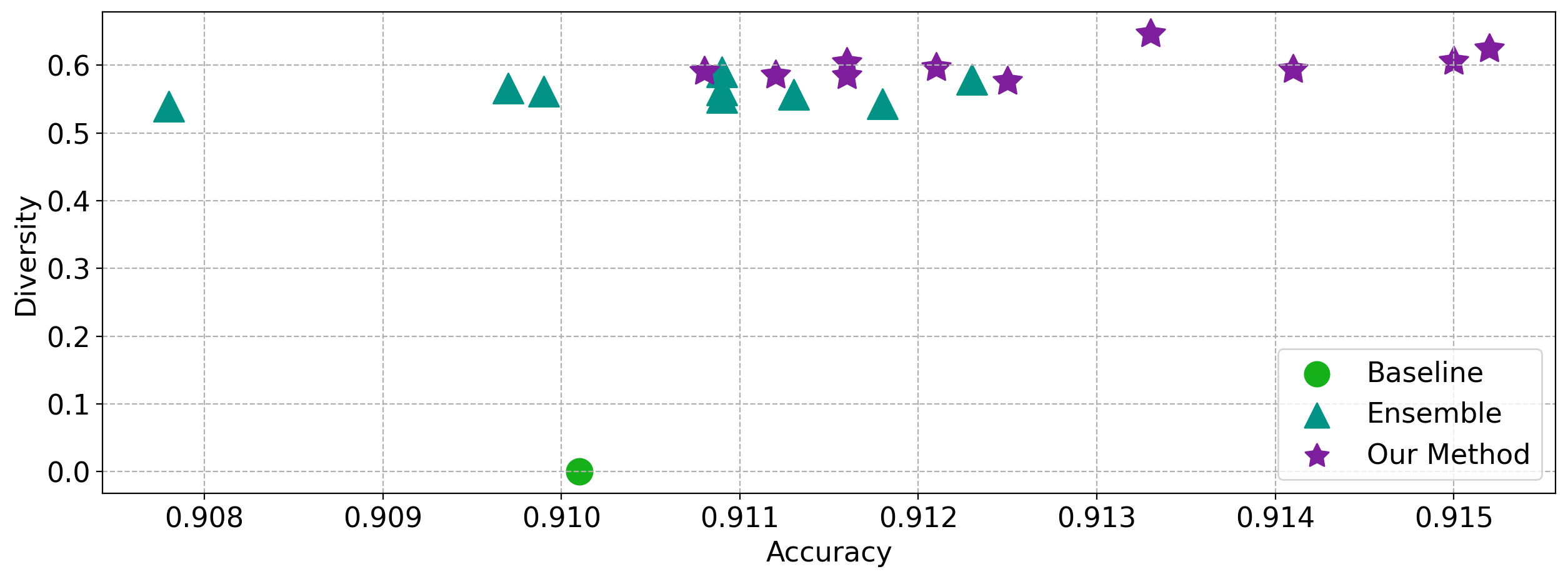}
  \vspace{-10pt}
  \caption{\small \textbf{Diversity Analysis}: prediction diversity disagreement vs. achieved accuracy on test set of CIFAR-10. Diversity analysis encompasses the comparative assessment of two distinct models that have been trained on test datasets, with a focus on quantifying the dissimilarity in their respective predictions. This evaluation entails computing the fraction of test data points for which predictions of models disagree, the diversity, and normalizing it by the model's error rate. Our method with $10$ subnetworks is on par with the deep self-supervised ensemble with $10$ members in terms of both accuracy and diversity disagreement. Models in the top right corner are better.}
\label{fig:diversity}
  \vspace{-8pt}
\end{figure}


\subsection{Empirical Analysis of Diversity}
Diversity of ensemble members is an important feature for powerful model ensembles and reflects the degree of independence among its members~\cite{zhang2012ensemble,ortega2022diversity}.
We follow~\citet{fort2019deep} to quantify the diversities among the ensemble of sub-networks. Specifically, we report the diversities in terms of \emph{disagreement score} between the members' predictive distributions and a baseline.
Diversity disagreement is defined as \textit{distance disagreement} divided by $1 -$ \textit{accuracy}, where the distance disagreement between two classification models $h_i$ and $h_j$ is calculated as $\frac{1}{N} \sum_{k=1}^N  \big[h_i(\mathbf{x}_k)\neq h_j(\mathbf{x}_k)\big],$ with $N$ denoting the number of samples. 
Figure~\ref{fig:diversity} compares the diversity disagreement between our method with $10$-sub-networks, a deep ensemble with $10$ members, and the single-network baseline. The results clearly indicate that our proposed method achieves comparable results with deep self-supervised ensembles in terms of both accuracy and diversity disagreement.


\subsection{Computational Cost and Efficiency Analysis}\label{sec:method:cost}
We analyze the efficiency of our proposed method in Table.~\ref{table:computationalcost}. SSL-Ensemble increases memory and computational requirements compared to the baseline by 200\% and 900\% for 3 and 10 members, respectively. The increase in the number of parameters is 32\% and 143\%, and the increase in computational requirement is $\sim 0-6\%$ for our method. A more detailed description of the relative cost and the reason for the difference between the increase in memory and the computational requirements of our method is provided in Appendix (see Computation Cost Analysis).

\vspace{-10pt}
\begin{table}[H] 
\centering \footnotesize
\caption{Computational cost in 4 DGX-A100 40G GPUs (PyTorch) on CIFAR 10.} 
\label{table:computationalcost}
  \scalebox{0.76}{
\begin{tabular}{lrrrr}
\toprule
Method          & Members & Parameters(M)    & Memory / GPU & Time / 800-ep. \\ \midrule
Baseline (SSL)  &  1      & 28          & 9 G         & 3.6 (h) \\
SSL-Ensemble    &  3      & 3$\times$28 & 3$\times$9 G       & 3$\times$ 3.6 (h)  \\
SSL-Ensemble    &  10     & 10$\times$28& 10$\times$9 G      & 10$\times$3.6 (h)  \\
Our method      &  3      & 37          & 9.2 G       & 3.6 (h) \\
Our method      &  10     & 68.1        & 10 G        & 3.8 (h) \\
\bottomrule
\end{tabular}
}
\end{table}

\section{Experimental Setup}
We perform several experiments with a variety of self-supervised methods to examine our hypothesis for robustness during both pretext-task learning and downstream tasks (fine-tuning). 

\noindent\textbf{Deep self-supervised network architecture ~}
Our proposed approach builds on two recent popular self-supervised models in computer vision: i) \textbf{SimCLR}~\citep{chen2020simple} is a contrastive learning framework that learns representations by maximizing agreement on two different augmentations of the same image, employing a contrastive loss in the latent embedding space of a convolutional network architecture (e.g., ResNet-50~\citep{he2016deep}), and
ii) \textbf{DINO}~\citep{caron2021emerging} is a self-distillation framework in which a student vision transformer (ViT; \citep{dosovitskiy_2021_ImageWorth16x16}) learns to predict global features from local image patches supervised by the cross-entropy loss from a momentum teacher ViT’s embeddings. 
Furthermore, we study the impact of our approach in NLP and modify \textbf{SCD} \citep{klein2022scd}, which applies the bidirectional training of transformers to language modeling. Here, the objective is self-supervised contrastive divergence loss.   
Lastly, we examine our approach on \textbf{Self-GenomeNet}~\citep{gunduz2021self}, a contrastive self-supervised learning algorithm for learning representations of genome sequences. 
More detailed descriptions of the employed configurations are provided in Appendix (see Implementation Details) 

\noindent\textbf{Deep independent sub-networks ~} 
We implement $M$ independent sub-networks on top of the encoder, for which many possible architectures are conceivable. For our experiments on computer vision datasets, we consider an ensemble of sub-network architecture where each network includes a multi-layer perceptron (MLP) with two layers of 2048 and 128 neurons, respectively, with ReLU as a non-linearity and followed by batch normalization \citep{ioffe_2017_BatchRenormalizationReducing}. Each sub-network has its own independent set of weights and learning parameters. For the NLP dataset, the projector MLP contains three layers of 4096 neurons each, also using ReLU activation's as well as batch normalization. For the genomics dataset, our ensemble of sub-networks includes one fully connected layer with an embedding size of 256.

\noindent\textbf{Optimization ~}
For all experiments on image datasets based on DINO and SimCLR, we follow the suggested hyperparameters and configurations by the paper~\citep{caron2021emerging,chen2020simple}. Implementation details for pretraining with DINO on the 1000-classes ImagetNet dataset without labels are as follows: coefficients $\epsilon$, $\alpha$, and $\lambda$ are respectively set to $0.0001,0.15,$ and $2$ in Eq.\ref{eq:loss:var}, \ref{eq:loss:div}, and \ref{eq:loss:total}. We provide more details in ablation studies (Section~\ref{ablation}) on the number of sub-networks and the coefficients $\lambda$ and $\alpha$ used in the loss function. The encoder network $f_{\bm{\theta}}$ is either a ResNet-50~\citep{he2016deep} with 2048 output units when the baseline is SimCLR~\citep{chen2020simple} or ViT-s~\citep{dosovitskiy2020image} with 384 output units when the baseline is DINO~\citep{caron2021emerging}. The best prediction and calibration performance is achieved when the number of sub-networks is 5. We followed the training protocol and settings suggested by~\citep{caron2021emerging}.

\noindent\textbf{Datasets ~}
We use the following datasets in our experiments: \textbf{CIFAR-10/100}~\citep{krizhevsky2009learning} are subsets of the tiny images dataset. Both datasets include 50,000 images for training and 10,000 validation images of size $32\times32$ with 10 and 100 classes, respectively. \textbf{SVH}~\citep{netzer2011reading} is a digit classification benchmark dataset that contains 600,000 $32\times32$ RGB images of printed digits (from 0 to 9) cropped from pictures of house number plates. \textbf{ImageNet}~\citep{deng2009imagenet}, contains 1,000 classes, with 1.28 million training images and 50,000 validation images. For the NLP task, we train on a dataset of 1 million randomly sampled sentences from \textbf{Wikipedia articles} \citep{wikipediadataset} and evaluate our models on 7 different semantic textual similarity datasets from the SentEval benchmark suite \citep{conneau2018senteval}: \textbf{MR} (movie reviews), \textbf{CR} (product reviews), \textbf{SUBJ} (subjectivity status), \textbf{MPQA} (opinion-polarity), \textbf{SST-2} (sentiment analysis), \textbf{TREC} (question-type classification), and \textbf{MRPC} (paraphrase detection). The \textbf{T6SS} effector protein dataset is a public real-world bacteria dataset (SecReT6, \citep{li2015secret6}) with actual label scarcity. The sequence length of the genome sample is 1000nt in all experiments.

\noindent\textbf{Tasks ~} 
We examine and benchmark a model’s performance on different tasks considering evaluation protocols by self-supervised learning~\citep{chen2020simple} and Plex's benchmarking tasks~\citep{tran2022plex}. Specifically, we evaluate our model on the basis of \textbf{uncertainty-aware IND generalization}, \textbf{OOD detection}, \textbf{semi-supervised learning}, \textbf{corrupted dataset evaluation} (see Section~\ref{sec:results}), and \textbf{transfer learning to other datasets and tasks} (see Appendix: Transfer to Other Tasks and Datasets )

\noindent\textbf{Evaluation metrics ~} 
We report prediction/calibration performance with the following metrics, where upward arrows indicate that higher values are desirable, \textit{et vice versa}. 
\textbf{Top-1 accuracy $\uparrow$}: share of test observations for which the correct class is predicted.
\textbf{AUROC} $\uparrow$: area under the ROC curve arising from different combinations of false-positive and false-negative rates (here: with positive and negative classes referring to being in and out of distribution, respectively) for a gradually increasing classification threshold.
\textbf{Negative log-likelihood (NLL)} $\downarrow$: negative log-likelihood of test observations under the estimated parameters.
\textbf{Expected calibration error (ECE)};\citep{naeini_2015_ece} $\downarrow$: mean absolute difference between accuracy and confidence (highest posterior probability among predicted classes) across equally-spaced confidence bins, weighted by relative number of samples per bin.
\textbf{Thresholded adaptive calibration error (TACE)}; \citep{nixon_2019_MeasuringCalibrationDeep}) $\downarrow$: modified ECE with bins of equal sample size, rather than equal interval width, and omitting predictions with posterior probabilities falling below a certain threshold (here: 0.01) that often dominate the calibration in tasks with many classes.

\noindent\textbf{Compared methods ~}
We compare our method to the following contenders.
\textbf{Baseline:} self-supervised architectures (i.e., SimCLR, DINO, SCD, or Self-GenomeNet, depending on the task). \textbf{SSL-Ensemble}: deep ensemble comprising a multiple of the aforementioned baseline networks. \textbf{Monte Carlo (MC) dropout}: \citep{gal2016dropout} baseline networks with dropout regularization applied during pretraining of baseline encoder. \textbf{BatchEnsemble}: baseline encoder with BatchEnsemble applied during pretraining.



\section{Results and Discussion}\label{sec:results} 

\textbf{In-distribution generalization ~}~\label{sec:results:ind}
IND generalization (or \textit{prediction calibration}) quantifies how well model confidence aligns with model accuracy. We perform several experiments on small and large image datasets as well as the genomics sequence dataset to evaluate and compare the predictive performance of our proposed model in IND generalization. Here, the base encoder $f_{\bm{\theta}}$ is frozen after unsupervised pretraining, and the model is trained on a supervised linear classifier. The linear classifier is a fully connected layer followed by softmax, which is placed on top of $f_{\bm{\theta}}$ after removing the ensemble of sub-networks. High predictive scores and low uncertainty scores are desired.

\begin{figure*}
  \centering
  \includegraphics[width=0.9\textwidth]{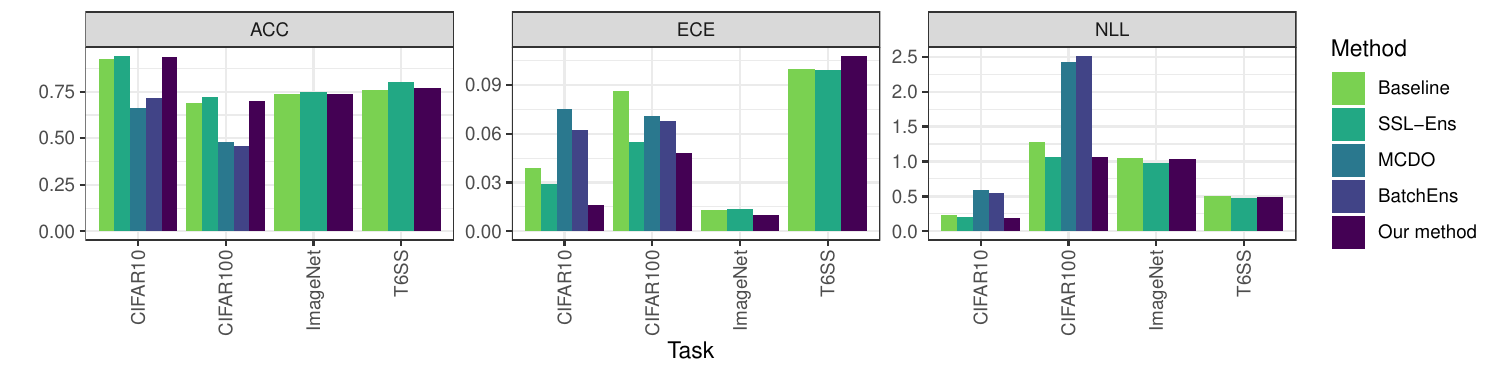}
  \caption{\textbf{IND generalization} in terms of (a) \textbf{Top-1 Accuracy} (b) \textbf{ECE} (c) \textbf{NLL} averaged over in-distribution on test samples of \textit{CIFAR-10/100, ImageNet, T6SS} datasets. Here, we compare our method with the ensemble of deep self-supervised networks (SSL-Ens), as well as the baseline. Detailed descriptions of IND generalization for each dataset and other competitors are presented in Appendix (see Additional Results) (Tables~\ref{table:calibration:cifar10}, \ref{table:calibration:cifar100}, \ref{table:calibration:imagenet}, and \ref{table:calibration:genome}).}
\label{fig:ind}
\vspace{-8pt}
\end{figure*}

Figure~\ref{fig:ind} illustrates the predictive probability of correctness for our model on CIFAR-10, CIFAR-100, ImageNet, and T6SS datasets in terms of Top-1 accuracy, ECE, and NLL, respectively. Based on Figure~\ref{fig:ind}, our method achieves better calibration (ECE and NLL) than the deep ensemble of self-supervised models.
The discrepancy in performance between our model and the deep ensemble can be explained by various factors, including differences in uncertainty modeling, complexity, and robustness. While the deep ensemble excels in top-1 accuracy, our model's superior ECE and NLL scores indicate better-calibrated and more reliable predictions, which are essential for safety-critical applications and decision-making under uncertainty. More detailed descriptions are provided in Appendix (see Additional Results) (Tables~\ref{table:calibration:cifar10}, \ref{table:calibration:cifar100}, \ref{table:calibration:imagenet}, and \ref{table:calibration:genome}).

\begin{table}
\caption{ \small \textbf{OOD detection}. Results reported using AUROC show our method enhances the baseline up to 6\%.}
\label{table:ood}
  \centering
  \scalebox{0.78}{
  \begin{tabular}{lcccccc}
   \toprule
       IND    &  OOD  & Baseline  & SSL-Ensemble & Our method  \\
   \midrule
    \multirow{4}{6em}{CIFAR-100} & SVHN & 84.22 & 84.95 & \textbf{88.00} \\ 
                                & Uniform & 91.65 & 90.53 & \textbf{97.57} \\ 
                                & Gaussian & 90.00 & 89.42 & \textbf{94.10} \\   
                                & CIFAR-10 & 74.71 & 74.80 & \textbf{75.18} \\ 
    \midrule                            
    \multirow{4}{6em}{CIFAR-10} & SVHN & 95.03 & 96.68 & \textbf{97.07}\\ 
                                & Uniform & 96.73 & 91.64 & \textbf{99.05}\\ 
                                & Gaussian & 96.39 & 93.24 & \textbf{99.24} \\   
                                & CIFAR-100 & 91.79 & 91.59 & \textbf{91.87} \\     
    \bottomrule
  \end{tabular}
  }
\end{table}
\textbf{Out-of-distribution detection ~} 
OOD detection shows how well a model can recognize test samples from the classes that are unseen during training~\citep{geng2020recent}. We perform several experiments to compare the model generalization from IND to OOD datasets and to predict the uncertainty of the models on OOD datasets. 
Evaluation is performed directly after unsupervised pretraining without a fine-tuning step. Table~\ref{table:ood} shows the AUROC on different OOD sets for our model, baseline, and deep self-supervised ensemble. Our approach improves overall compared to other methods.


\begin{table}
\caption{\small \textbf{Semi-supervised evaluation}: Top-1 accuracy (ACC), ECE, and NLL for semi-supervised CIFAR-10/100 classification using 1\% and 10\% training examples.}
\label{table:semisupervised:cifar10}
  \centering
  \scalebox{0.5}{
  \begin{tabular}{l|ccc|ccc|ccc|ccc}
   \toprule
       Method    & \multicolumn{3}{c}{ CIFAR-10 (1\%)} & \multicolumn{3}{c}{CIFAR-10 (10\%)} & \multicolumn{3}{|c}{ CIFAR-100 (1\%)} & \multicolumn{3}{c}{CIFAR-100 (10\%)}\\
        \midrule
                & ACC & ECE & NLL & ACC & ECE & NLL & ACC & ECE & NLL & ACC & ECE & NLL  \\
    \midrule
    Baseline      & 89.1 & 0.075 & 0.364 & 91.1 & 0.039 & 0.274 & 56.2 & 0.097 & 2.01 & 59.5 & 0.086 & 1.79 \\   
    SSL-Ensemble  & 90.1 & 0.056 & 0.334 & 92.2 & 0.050 & 0.257 & 59.7 & 0.081 & 1.86 & 62.6 & 0.053 & 1.48 \\   
    Our method    & 90.4 & 0.018 & 0.296 &  92.6 & 0.016 & 0.249 & 59.3 & 0.060 & 1.71 & 62.4 & 0.042 & 1.56\\   
    \bottomrule
  \end{tabular}
  }
\end{table}

\textbf{Semi-supervised evaluation ~}
We explore and compare the performance of our proposed method in the low-data regime. Again, the encoder $f_{\bm{\theta}}$ is frozen after self-supervised pretraining, and the model is trained on a supervised linear classifier using 1\% and 10\% of the dataset. The linear classifier is a fully connected layer followed by softmax. Table~\ref{table:semisupervised:cifar10} shows the result in terms of top-1 accuracy, ECE, and NLL. The results indicate that our method outperforms other methods in the low-data regime -- in terms of calibration.


\begin{figure}
\centering
    \includegraphics[width=0.5\textwidth]{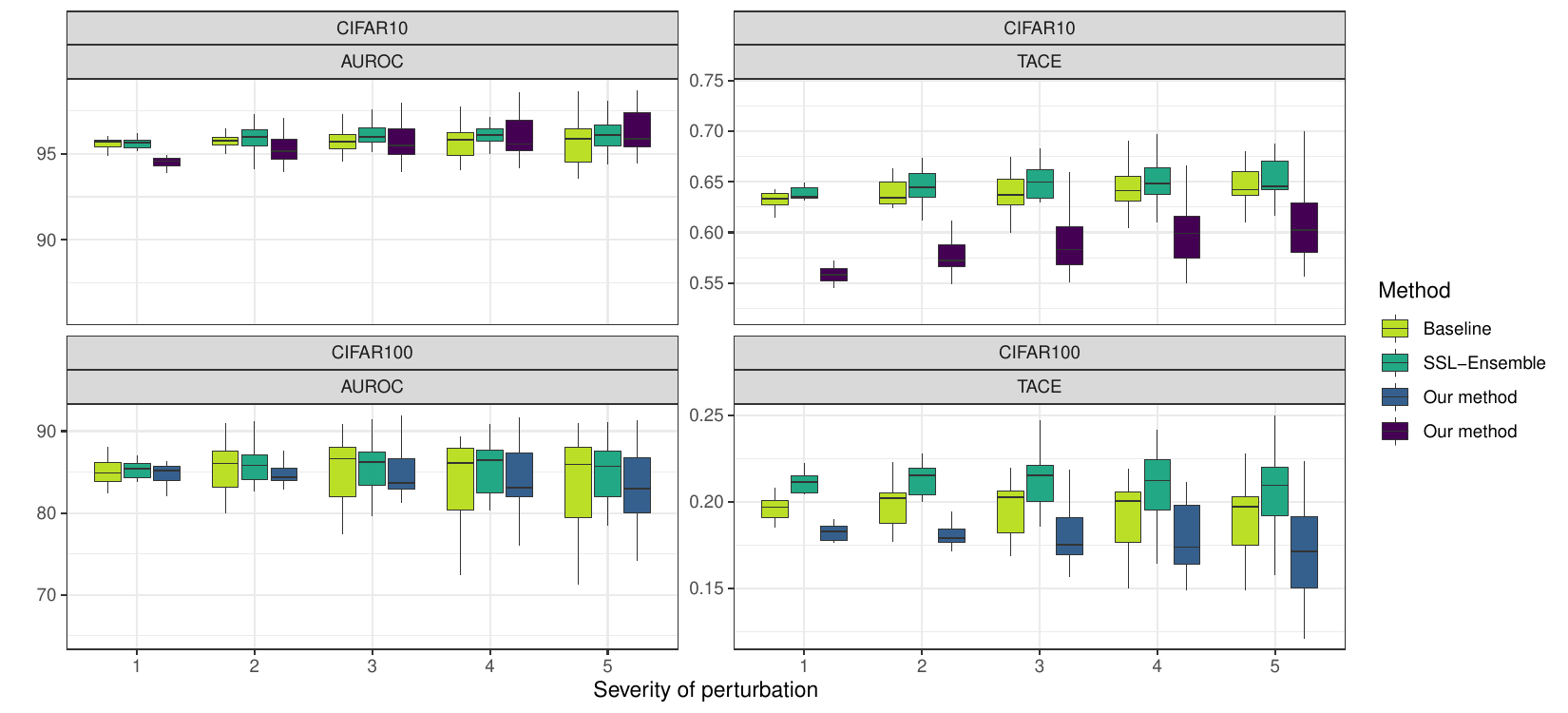}
    \caption{Performance under \textbf{dataset corruption} (CIFAR-10/100 with five levels of increasing perturbation), evaluation in terms of AUROC and TACE for several types of corruption (vertical spread).} 
\label{fig:covariateshift}
\vspace{-10pt}
\end{figure}

\textbf{Corrupted dataset evaluation ~}
Another important component of model robustness is its ability to make accurate predictions when the test data distribution changes. Here, we evaluate model robustness under \emph{covariate shift}. We employ a configuration similar to the one found in \citep{tran2022plex}. Figure~\ref{fig:covariateshift} summarizes the improved performance across metrics of interest. The results confirm that our method outperforms the baseline and achieves comparable predictive performance as a deep self-supervised ensemble -- both in terms of calibration (TACE) and AUROC.  
\section{Ablation Study} \label{ablation} 

\begin{figure}[!t]
  \centering
  \includegraphics[width=0.45\textwidth]{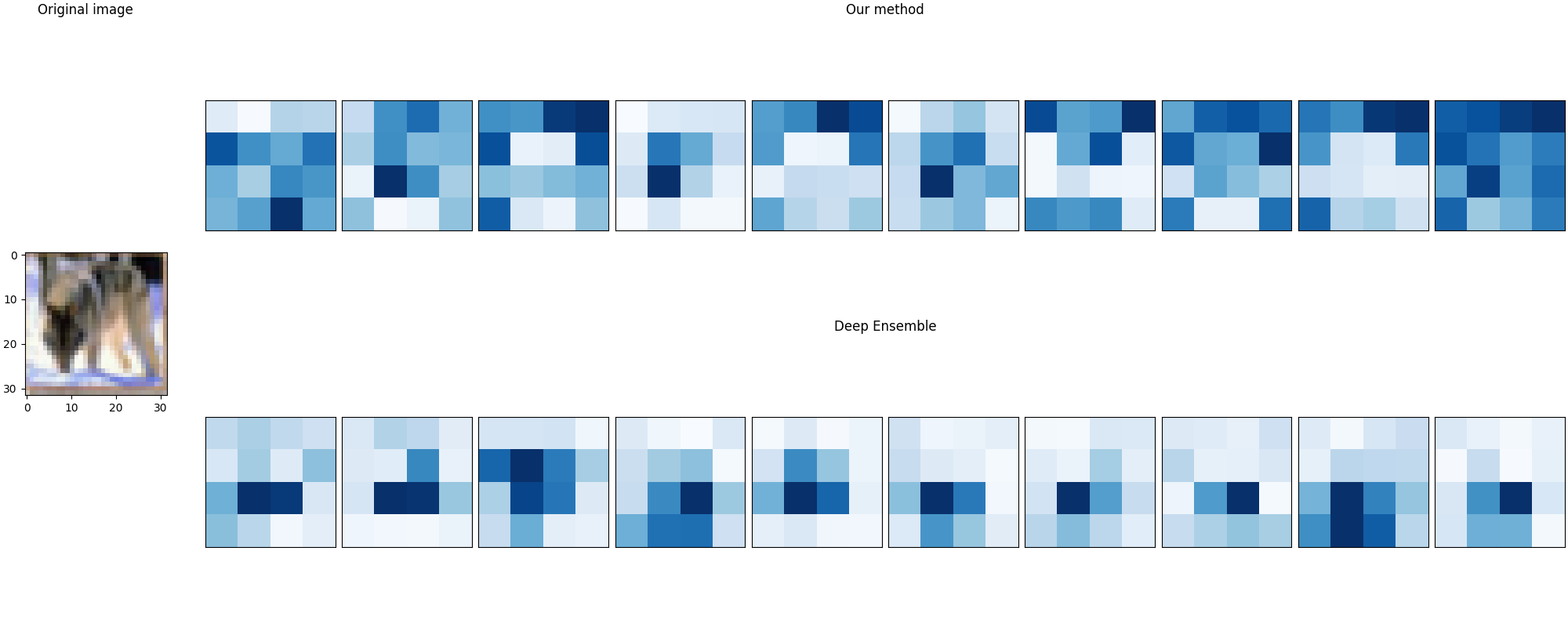}
  \caption{We compare the feature diversity for different subnetworks and ensemble members. The top images are for different sub-networks, and the bottom images are for different ensemble members. We used Grad-CAM \citep{selvaraju2017grad} for visualization.} 
\label{fig:ablation: diversity}
\vspace{-10pt}
\end{figure}

In order to build intuition around the behavior and the observed performance of the proposed method, we further investigate the following aspects of our approach in multiple ablation studies exploring: (1) the number $M$ of sub-networks, (2) the role of each component of the proposed loss, and (3) analysis of diversity with visualization of the gradients of subnetworks. We also present more results on (4) the impact of our approach during pretraining vs. at the finetuning step, (5) the size of sub-networks, and (6) the impact of model parameters in the Appendix (see Additional Ablation Analysis) .

\begin{figure}[!t]
  \centering
  \subfloat[]{\includegraphics[width=0.5\textwidth]{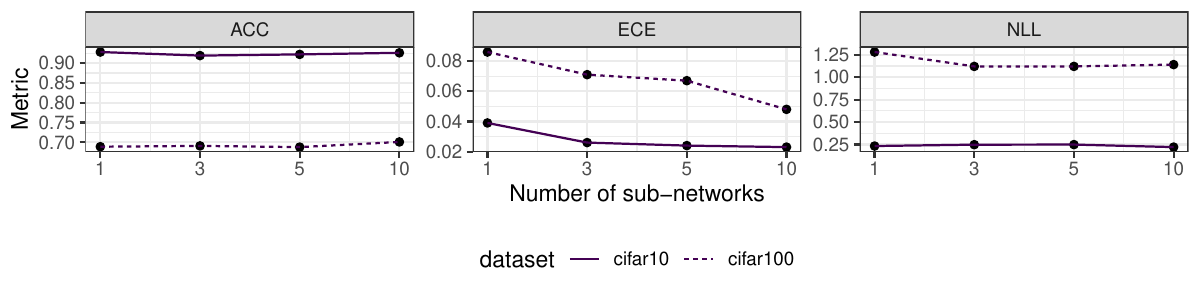} \label{fig:Mheads}} \\
  \centering
  \subfloat[]{\includegraphics[width=0.45\textwidth]{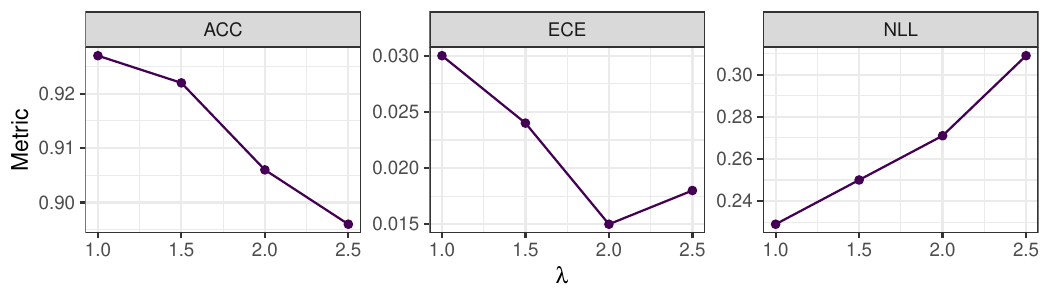}\label{fig:alpha}} \\
  \centering
  \subfloat[]{\includegraphics[width=0.45\textwidth]{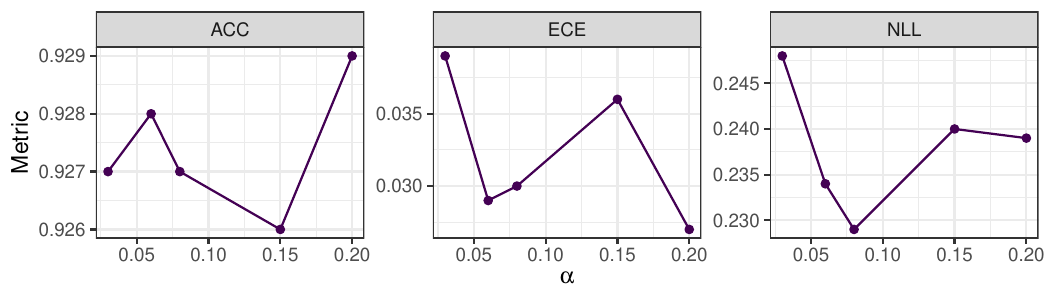}\label{fig:lambda}}
  \caption{\small Ablation study on number of $M$ sub-networks (a), hyperparameters of our proposed loss \small (b) \textbf{$\lambda$} and (c) \textbf{$\alpha$}.}
\label{fig:ablation:loss}
\end{figure}

\textbf{Number of sub-networks ~} 
We train $M$ individual deep neural networks on top of the representation layer. The networks receive the same inputs but are parameterized with different weights and biases. Here, we provide more details regarding our experiments on IND generalization by considering varying $M$. 
Fig.~\ref{fig:Mheads} compares the performance in terms of top-1 accuracy, ECE, and NLL for CIFAR-10 and CIFAR-100. 
Based on the quantitative results depicted in Fig.~\ref{fig:Mheads}, the predictive performance improves in both datasets when increasing the number of sub-networks ($M$) until a certain point. 
For example, in the case of CIFAR-10, when $M = 3$, our performance is $91.9\%$; increasing $M$ to 10 levels top-1 accuracy up to $92.6\%$, while the ECE and NLL decrease from $0.026$ and $0.249$ to $0.023$ and $0.222$, respectively. These findings underline that training our sub-networks with a suitable number of heads can lead to a better representation of the data and better calibration. 
Recently \citep{wen2022mechanism, tian2021understanding} provided a theoretical statement as well as experimental results that projection heads help with faster convergence.

\textbf{Analysis of loss ~}
The total loss (Eq.~\ref{eq:loss:total}) is calculated by the combination of self-supervised loss (Eq.~\ref{eq:loss:ssl}) and diversity loss (Eq.~\ref{eq:loss:div}), where the mean value of the embeddings across the ensemble of sub-networks is fed to the self-supervised loss, and the corresponding standard deviation is used for the diversity loss. 
First, we note that the use of our diversity regularizer indeed improves calibration and provides better uncertainty prediction.
The results in Fig.~\ref{fig:ind} show the impact of our loss function in relation to the baseline. By comparing the first and fifth rows of Table~\ref{table:calibration:cifar10}, it can be inferred that our proposed loss function results in a much lower ECE ($0.016$) than the network trained by SimCLR (baseline) with $0.039$ on the CIFAR-10 dataset. Similarly, the first and third rows of Table \ref{table:calibration:imagenet} compare the predictive probability of correctness of DINO (baseline) and our model on ImageNet.  

Second, we explore different hyperparameter configurations to find the optimal values for $\alpha$ and $\lambda$ in Fig. ~\ref{fig:alpha}, \ref{fig:lambda}. 
Note that, in practice, $\alpha$ and  
$\lambda$ must be optimized jointly. The best top-1 accuracy in our case is achieved when $\alpha$ and $\lambda$ are set to 0.08 and 1.5, respectively, on the CIFAR-10 dataset.

\textbf{Analysis of diversity ~}
In addition to quantitative results for diversity analysis provided in Figure \ref{fig:diversity}, we visualize the activation map for the last convolution layer in the encoder for each ensemble member and each subnetwork to motivate the effect of subnetworks on the encoder. As illustrated in Fig.~\ref{fig:ablation: diversity}, different subnetworks have more feature diversity compared to the deep ensemble as we expected.

\section{Conclusion} \label{conclusion}

In this paper, we presented a novel diversified ensemble of self-supervised framework. We achieved high predictive performance and good calibration using a simple yet effective idea -- an ensemble of independent sub-networks. We introduced a new loss function to encourage diversity among different sub-networks.
It is straightforward to add our method to many existing self-supervised learning frameworks during pretraining. Our extensive experimental results show that our proposed method outperforms, or is on par with, an ensemble of self-supervised baseline methods in many different experimental settings.

\newpage
{
\small
\bibliography{aaai24}
}

\newpage

\section{Implementation Details}~\label{sec:app}

\subsection{Computation Cost Analysis}

\begin{figure}[H]
\centering
    \includegraphics[width=0.5\textwidth]{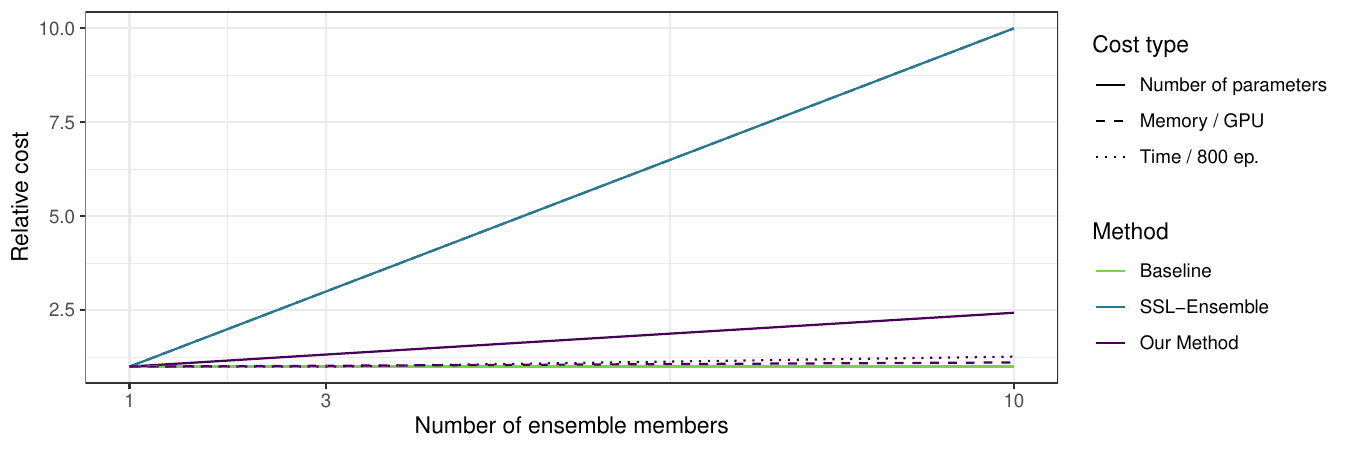}
    \caption{The test time cost (purple, dotted) and memory cost (purple, dashed) of our model w.r.t. the ensemble size. The figures are relative to the cost incurred by a single model (green). The inference time cost and memory cost of a deep self-supervised ensemble are plotted in blue.} 
\label{fig:relative_cost}
\end{figure}

Figure~\ref{fig:relative_cost} illustrates relative computation cost -- as compared to the baseline -- in terms of the number of parameters, computation time, and memory required between our model and a deep self-supervised ensemble. 

\subsection{Computational Cost Analysis}

As we mentioned in Section Method 
the increase in the number of parameters is 32\% and 143\%, and the increase in computational requirement is negligible and 6\% for our method compared to the baseline when there exist 3 and 10 ensemble members, respectively. We would like to explain the reason as follows:

While the encoder networks used in the baseline methods (and our method) contain many convolutional layers, the additional parameters introduced by our method are in the projection head, and they are a few linear layers. Although these additional linear layers increase the number of parameters to some extent, the computational burden introduced by them is much more limited compared to the convolutional layers that exist in both baselines and our method. That is because convolutional layers typically contain fewer parameters compared to fully connected layers due to parameter-sharing but have a much higher computational burden since their outputs are evaluated over the whole image. A toy example to understand this would be the comparison of the two settings below:

\begin{enumerate}
  \item Consider a 100 $\times$ 100 image fed into a convolutional filter with (kernel size= 3 $\times$ 3, stride= 1 $\times$ 1, padding= ”same”, bias= False). The convolutional filter has 9 parameters but needs to do 100 $\times$ 100 $\times$ 9 = 90000 multiplications to evaluate its output. 

\item Consider a vector of 1000 that is fed into a fully connected layer (without bias) to produce 1 output value. The fully connected layer contains 1000 parameters and the number of multiplications needed to evaluate its output is also 1000. 
\end{enumerate}

Comparing these two settings, the convolutional layer needs $\sim$100 times more computational burden (convolutional: 90000 vs dense: 1000 multiplications) to evaluate its outcome, although it has $\sim$100 times fewer parameters (convolutional: 9 vs dense: 1000) compared to the fully connected layer.

Similarly, the increase in memory requirements is low for our method compared to the SSL-Ensemble, but the increase in computational requirements is much lower and even negligable.

\subsection{Data Augmentation for Computer Vision Datasets}~\label{sec:app:aug}

We define a random transformation function $\bm T$ that applies a combination of crop, horizontal flip, color jitter, and grayscale. Similar to~\cite{chen2020simple}, we perform crops with a random size from $0.2$ to $1.0$ of the original area and a random aspect ratio from $3/4$ to $4/3$ of the original aspect ratio. We also apply horizontal mirroring with a probability of $0.5$. Then, we apply grayscale with a probability of $0.2$ as well as color jittering with a probability of $0.8$ and a configuration of $(0.4, 0.4, 0.4, 0.1)$. However, for ImageNet, we define augmentation based on the original DINO from their official repository. In all experiments, at the testing phase, we apply only resize and center crop.

\subsection{Hyperparameters for Self-supervised Network Architectures} \label{sec:app:arch}

\noindent\textbf{SimCLR}~\citep{chen2020simple}: 
we use ResNet-50 as a backbone, a loss temperature of 0.07, batch size 512, and a cosine-annealing learning rate scheduler. The embedding size is 2048, and we train for 800 epochs during pretraining.
\noindent\textbf{DINO}~\citep{caron2021emerging}: 
we use ViT-small as a backbone, patch size 16, batch size 1024, and a cosine-annealing learning rate scheduler. The embedding size is 384/1536, and we train for 100 epochs during pretraining.



\section{Additional Results}

\subsection{Robustness of representation: IND- Generalization}~\label{sec:app:ind}

Tables~\ref{table:calibration:cifar10}, \ref{table:calibration:cifar100}, and \ref{table:calibration:imagenet} present results for the predictive performance and calibration of our model on CIFAR-10, CIFAR-100, and ImageNet respectively. Based on Table~\ref{table:calibration:cifar10}, our method achieves better calibration than the deep ensemble of self-supervised networks, MC-Dropout, and BatchEnsemble, with significant margins at large ensemble sizes. In order to have multiple batches for BatchEnsemble, we decreased the initial batch size because of memory, so we ended up with a smaller batch size to which the self-supervised model (i.e., SimCLR) is sensitive. Also, each time we have more positive samples than the original.

In the case of dropouts, we again face the same problem with positive and negative samples. Dropouts also count as data regularization, but when applied randomly to all data in contrastive learning, it degrades the idea of positive and negative. For example, in NLP, dropouts are used to produce different augmentations.


\begin{table*}
\caption{\textbf{ IND Generalization}: Top-1 accuracy, ECE and NLL averaged over in-distribution on test samples of the \textbf{CIFAR-10} dataset over three random seeds. The best score for each metric is shown in \textbf{bold}, and the second-best is \underline{underlined}.}
\label{table:calibration:cifar10}
  \centering
  \scalebox{0.78}{
  \begin{tabular}{l|ccc|ccc|ccc}
   \toprule
       Method    & \multicolumn{3}{c}{Top-1 Acc (\%)  ($\uparrow$)} & \multicolumn{3}{|c}{ECE  ($\downarrow$)}  & 
       \multicolumn{3}{|c}{NLL  ($\downarrow$)}                  \\
        \midrule
      \# member ($M$)  & 3 & 5 & 10 & 3 & 5 & 10 & 3 & 5 & 10  \\
    \midrule
    Baseline & \multicolumn{3}{c|}{92.8 $\pm$ 0.4} & \multicolumn{3}{c}{0.039 $\pm$ 0.002} & \multicolumn{3}{c}{0.233 $\pm$ 0.011} \\
    SSL-Ensemble & $92.8 \pm 0.1$ & $ 93.0 \pm 0.2$ & \textbf{94.2 $\pm$ 0.3} & $0.043 \pm 0.02$ & $ 0.033 \pm 0.01$ & \underline{0.029 $\pm$ 0.02} & $0.221 \pm 0.011$ & $0.226 \pm 0.009$ & \underline{$0.199 \pm 0.004$} \\ 
    MC Dropout& $65.7 \pm 0.2$ & $ 66.3 \pm 0.2$ & $66.4 \pm 0.2$ & $0.083 \pm 0.014$ & $0.077 \pm 0.009$ & $0.075 \pm 0.005$ & $ 0.66 \pm 0.012$ & $0.637 \pm 0.002$ & $0.593 \pm 0.006$  \\ 
    BatchEnsemble & $69.1 \pm 04$ & $ 72.1 \pm 0.3$ & $ 71.9 \pm 0.2$ & $ 0.064 \pm 0.011$ & $0.061 \pm 0.008$ & $0.062 \pm 0.005$ & $0.613 \pm xx$ & $0.58 \pm 0.007$ & $0.551 \pm 0.004$ \\
    Our method & $92.6 \pm 0.2$ & $ 92.9 \pm 0.1$ & \underline{$93.6 \pm 0.1$} & $0.021 \pm 0.004$ & $ 0.019 \pm 0.002$ & \textbf{0.016 $\pm$ 0.001} & $0.241 \pm 0.010$ & $0.221 \pm 0.005$ & \textbf{0.193 $\pm$ 0.003} \\       
    \bottomrule
  \end{tabular}
  }
\end{table*}


\begin{table*}
\caption{\textbf{ IND Generalization}: Top-1 accuracy, ECE and NLL averaged over in-distribution on test samples of the \textbf{CIFAR-100} dataset over three random seeds. The best score for each metric is shown in \textbf{bold}, and the second-best is \underline{underlined}.}
\label{table:calibration:cifar100}
  \centering
  \scalebox{0.77}{
  \begin{tabular}{l|ccc|ccc|ccc}
   \toprule
       Method    & \multicolumn{3}{c}{Top-1 Acc (\%)  ($\uparrow$)} & \multicolumn{3}{|c}{ECE  ($\downarrow$)}  & 
       \multicolumn{3}{|c}{NLL  ($\downarrow$)}                  \\
        \midrule
      \# member ($M$)  & 3 & 5 & 10 & 3 & 5 & 10 & 3 & 5 & 10  \\
    \midrule
    Baseline & \multicolumn{3}{c|}{68.9 $\pm$ 0.3} & \multicolumn{3}{c}{0.086 $\pm$ 0.014} & \multicolumn{3}{c}{1.28 $\pm$ 0.05} \\
    SSL-Ensemble & $70.6 \pm 0.12$ & $ 71.4 \pm 0.5$ & \textbf{72.0 $\pm$ 0.2} & $0.12 \pm 0.01$ & $ 0.122 \pm 0.01$ & $0.119 \pm 0.04$ & $1.09 \pm 0.01$ & $1.12 \pm 0.01$ & \underline{$1.06 \pm 0.02$} \\ 
    MC Dropout& $46.3 \pm 0.1$ & $ 45.2 \pm 0.4$ & $48.2 \pm 0.1$ & $0.077 \pm 0.012$ & $0.081 \pm 0.002$ & $0.071 \pm 0.002$ & $ 2.66 \pm 0.11$ & $2.37 \pm 0.02$ & $2.43 \pm 0.06$  \\ 
    BatchEnsemble & $44.1 \pm 03$ & $ 45.2 \pm 0.2$ & $ 46.1 \pm 0.1$ & $ 0.073 \pm 0.01$ & $0.071 \pm 0.08$ & \underline{0.068 $\pm$ 0.001} & $2.43 \pm 0.03$ & $2.64 \pm 0.007$ & $2.51 \pm 0.004$ \\
    Our method & $67.7 \pm 0.1$ & 68.8 $\pm$ 0.1 & \underline{$70.1 \pm 0.0$} & $0.067 \pm 0.001$ & $ 0.063 \pm 0.001$ & \textbf{0.048 $\pm$ 0.000} & $0.114 \pm 0.005$ & $0.116 \pm 0.0002$ & \textbf{1.06 $\pm$ 0.001} \\       
    \bottomrule
  \end{tabular}
  }
\end{table*}


\begin{table}[H]
\caption{\textbf{ IND Generalization}: Top-1 accuracy, ECE and NLL averaged over in-distribution on test samples of the \textbf{ImageNet} dataset over three random seeds. The best score for each metric is shown in \textbf{bold}, and the second-best is \underline{underlined}.}
\label{table:calibration:imagenet}
  \centering
  \scalebox{.8}{
  \begin{tabular}{l|c|c|c}
   \toprule
       Method    & \multicolumn{1}{c}{Top-1 Acc (\%)  ($\uparrow$)} & \multicolumn{1}{|c}{ECE  ($\downarrow$)}  & 
       \multicolumn{1}{|c}{NLL  ($\downarrow$)}  \\
    \midrule
    Baseline & 73.8 $\pm$ 0.3 & 0.013 $\pm$ 0.015 & 1.05 $\pm$ 0.01 \\
    SSL-Ensemble & \textbf{75.1 $\pm$ 0.1} & \underline{0.014 $\pm$ 0.000} & \textbf{0.98 $\pm$ 0.01} \\ 
    Our method & \underline{74.0 $\pm$ 0.0} & \textbf{0.010 $\pm$ 0.000} & \underline{1.03 $\pm$ 0.01} \\       
    \bottomrule
  \end{tabular}
  }
\end{table}


\begin{table}[H]
\caption{\textbf{IND Generalization}: Top-1 accuracy, ECE and NLL averaged over in-distribution on test samples of the \textbf{T6SS Identification} dataset over three random seeds. The best score for each metric is shown in \textbf{bold}, and the second-best is \underline{underlined}.}
\label{table:calibration:genome}
  \centering
  \scalebox{.8}{
  \begin{tabular}{l|c|c|c}
   \toprule
       Method    & \multicolumn{1}{c}{Top-1 Acc (\%)  ($\uparrow$)} & \multicolumn{1}{|c}{ECE  ($\downarrow$)}  & 
       \multicolumn{1}{|c}{NLL  ($\downarrow$)}  \\
    \midrule
    Baseline & 75.9 $\pm$ 2.0 & \underline{0.100 $\pm$ 0.006} & 0.502 $\pm$ 0.020 \\
    SSL-Ensemble & \textbf{80.2 $\pm$ 0.7} & \textbf{0.099 $\pm$ 0.014} & \textbf{0.471 $\pm$ 0.011} \\ 
    Our method & \underline{76.7 $\pm$ 2.3} & 0.108 $\pm$ 0.006 & \underline{0.492 $\pm$ 0.024}\\       
    \bottomrule
  \end{tabular}
  }
\end{table}

We also performed experiments on a dataset of 1-dimensional genomic sequences -- the T6SS identification of effector proteins-- to demonstrate that uncertainty-aware subnetworks can also be readily combined with existing models for 1-dimensional datasets and models. Based on Table~\ref{table:calibration:genome}, our method improves the accuracy and the calibration compared to the baseline. 


\subsection{Transfer to Other Tasks and Datasets} \label{sec:app:transfer}

\begin{figure*}
\centering
    \includegraphics[width=1.0\textwidth]{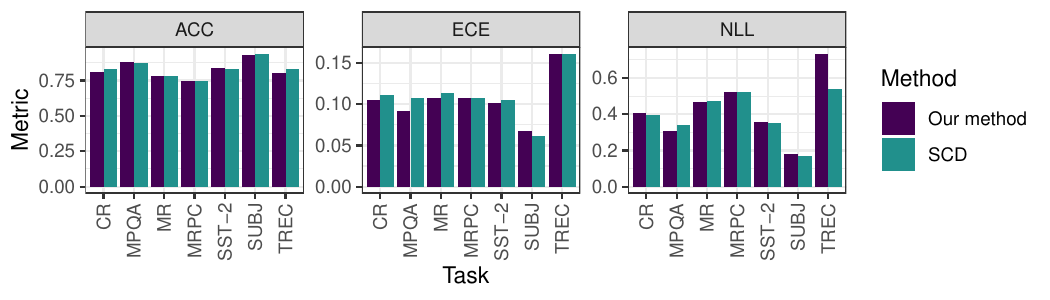}
    \caption{\textbf{Transfer to other dataset and tasks:} Comparision of  Sentence embedding performance on semantic textual similarity tasks.} 
\label{fig:transfer:nlp}
\end{figure*}

We further assess the generalization capacity of the learned representation on learning a new task in NLP. We train our model without any labels on a dataset of sentences from Wikipedia~\citep{wikipediadataset} and fine-tune the pretrained representation on seven different semantic textual similarity datasets from the SentEval benchmark suite \citep{conneau2018senteval}: \textbf{MR} (movie reviews), \textbf{CR} (product reviews), \textbf{SUBJ} (subjectivity status), \textbf{MPQA} (opinion-polarity), \textbf{SST-2} (sentiment analysis), \textbf{TREC} (question-type classification), and \textbf{MRPC} (paraphrase detection).
Then, we evaluate the test set of each dataset. Figure~\ref{fig:transfer:nlp} provides a comparison of the transfer learning performance of our self-supervised approach for different tasks. Our results in Figure~\ref{fig:transfer:nlp}  indicate that our approach performs comparably to or better than the baseline method.

We test the performance of the trained model on ImageNet~\citep{deng2009imagenet} on CIFAR-10~\cite{krizhevsky2009learning} dataset where the model is trained for 100 epochs.

\begin{table}[H]
\caption{\textbf{Transfer to other dataset}: Expected calibration error averaged over uncertainty-aware evaluation on CIFAR-10 datasets. }
\label{table:transfercv}
  \centering
  \scalebox{0.9}{
  \begin{tabular}{l|cccc}
   \toprule
    Method       & \multicolumn{1}{c}{ACC (\%)  ($\uparrow$)} & \multicolumn{1}{c}{ECE  ($\downarrow$)} & \multicolumn{1}{c}{NLL ($\downarrow$)} & \multicolumn{1}{c}{TACE  ($\downarrow$)} \\
    \midrule
    Baseline    & 73.5 & 0.038 & 0.78 & 0.20  \\
    Our method  & 73.9 & 0.030 & 0.75 & 0.18  \\
    \bottomrule
  \end{tabular}
  }
\end{table}

\section{Additional Ablation Analysis}

\textbf{Efficient ensemble of sub-networks at pretraining vs. finetuning}
We performed additional experiments to study the efficiency of proposed loss and independent sub-networks (InSub) i) during pretraining, ii) during finetuning, and iii) during both pretraining and finetuning. As shown in Table~\ref{table:finetunevspretrain}, pretraining with an ensemble of sub-networks is beneficial, and additional fine-tuning with multiple heads can further improve performance.

\begin{table*}
\caption{\textbf{Pretraining vs. Finetuning}: Expected calibration error averaged over uncertainty-aware evaluation on CIFAR-10 datasets. InSub refers to training with our proposed \underline{In}dependent \underline{Sub}networks }
\label{table:finetunevspretrain}
  \centering
  \scalebox{0.86}{
  \begin{tabular}{l|cccc}
   \toprule
    Method       & \multicolumn{1}{c}{ACC (\%)  ($\uparrow$)} & \multicolumn{1}{c}{ECE  ($\downarrow$)} & \multicolumn{1}{c}{NLL ($\downarrow$)} & \multicolumn{1}{c}{TACE  ($\downarrow$)} \\
    \midrule
    Baseline  & 92.5 & 0.039 & 0.238 & 0.133  \\
    Pretrain-InSub  & 92.6 & 0.032 & 0.226 & 0.131  \\
    Finetune-InSub  & 92.6 & 0.021 & 0.222 & 0.103 \\
    Pretrain-InSub + Finetune-InSub  & 92.8 & 0.023 & 0.227 & 0.115 \\
    \bottomrule
  \end{tabular}
  }
\end{table*}
\begin{table*}
\caption{\textbf{Sub-Network Size}: Expected calibration error averaged over uncertainty-aware evaluation on CIFAR-10 datasets. }
\label{table:subnet-size}
  \centering
  \scalebox{0.8}{
  \begin{tabular}{l|cccc}
   \toprule
    Method       & \multicolumn{1}{c}{ACC (\%)  ($\uparrow$)} & \multicolumn{1}{c}{ECE  ($\downarrow$)} & \multicolumn{1}{c}{NLL ($\downarrow$)} \\
    \midrule
    Our method with 5 sub-network  (100\%)  & 92.9 & 0.019 & 0.221   \\
    With 25 percent of sub-network size  & 92.3 & 0.026 & 0.231   \\
    With 50 percent of sub-network size  & 92.6 & 0.021 & 0.226   \\
    With 75 percent of sub-network size  & 92.6 & 0.019 & 0.221   \\
    \bottomrule
  \end{tabular}
  }
\end{table*}
\begin{table*}
\caption{\textbf{Large variant encoder}: Expected calibration error averaged over uncertainty-aware evaluation on CIFAR-10 datasets.}
\label{table:app:basline}
  \centering
  \scalebox{0.8}{
  \begin{tabular}{l|cccc}
   \toprule
    Method       & \multicolumn{1}{c}{ACC (\%)  ($\uparrow$)} & \multicolumn{1}{c}{ECE  ($\downarrow$)} & \multicolumn{1}{c}{NLL ($\downarrow$)} & \multicolumn{1}{c}{Number of parameters (M)} \\
    \midrule
    Our method with ResNet50 as a encoder with 5 sub-networks    & 92.9 & 0.019 & 0.221 & 45.79  \\
    Baseline with ResNet101 as a encoder  & 93.2 & 0.027 & 0.202 & 46.95  \\
    \bottomrule
  \end{tabular}
  }
\end{table*}

\begin{table*}
\caption{\textbf{Different encoder (medium size)}: Expected calibration error averaged over uncertainty-aware evaluation on CIFAR-10 datasets. }
\label{table:app:encoder}
  \centering
  \scalebox{0.8}{
  \begin{tabular}{l|cccc}
   \toprule
    Method       & \multicolumn{1}{c}{ACC (\%)  ($\uparrow$)} & \multicolumn{1}{c}{ECE  ($\downarrow$)} & \multicolumn{1}{c}{NLL ($\downarrow$)} & \multicolumn{1}{c}{Number of parameters (M)} \\
    \midrule
    Our method with ResNet34 as a encoder with 20 sub-networks    & 92.5 & 0.016 & 0.23 & 27.84  \\
    Baseline with ResNet50 as a encoder  & 92.8 & 0.039 & 0.233 & 27.89  \\
    \bottomrule
  \end{tabular}
  }
\end{table*}

\subsection{Analysis of Size of Sub-Networks}~\label{sec:app:ab:size}
We perform several experiments to study the different sizes of sub-network. As shown in Table~\ref{table:subnet-size}, the dimension of projection heads does not change the top-1 accuracy. Recent self-supervised models such as SimCLR \citep{chen2020simple}, BarlowTwins~\citep{zbontar2021barlow} also reach the same results with different projection head sizes.

\subsection{Impact of Model Parameters}~\label{sec:app:ab:baseline}

Our project aims to improve the predictive uncertainty of the baseline without losing predictive performance by mimicking the ensembles of self-supervised models with much lower computational costs. According to the results shown in Table~\ref{table:app:basline}, a bigger encoder can potentially improve the predictive performance, but it does not necessarily improve the predictive uncertainty of the results. We used ResNet101 as a baseline with more parameters in the encoder. To have a fair comparison, we compare it with our model with five heads. Our model performs better in ECE and NLL and has comparable accuracy.

Also, we used ResNet34 as a baseline with fewer parameters in the encoder with twenty heads and compared it with baseline ResNet50 with one head. According to results obtained in Table \ref{table:app:encoder}, our model performs better in terms of ECE and NLL and has on-par accuracy.

\subsection{Source Code}
Please find the source code in the supplementary material.

\section{Theoretical Supplement}

\subsection{Proof for Eq.~\ref{eq:derivative:loss}}~\label{sec:app:proof1}
\begin{equation}
    \begin{aligned}
    {\frac{\partial \left( \ell_{div} \right) }{ \partial \bm{z}_{k,\hat{m}, o}}} =     \tfrac{-1}{2} \underbrace{\left(\tfrac{1}{M-1}  \textstyle\sum_{m=1}^M (\bm{z}_{k, m, o} - \bar{\bm{z}}_{k, o})^2\right)}_{A} \cdot {\frac{\partial \left(\tfrac{1}{M-1}  \textstyle\sum_{m=1}^M (\bm{z}_{k, m, o} - \bar{\bm{z}}_{k, o})^2 \right) }{ \partial \bm{z}_{k,\hat{m},o}}} &=  \\ 
    \tfrac{-A}{2} \cdot \tfrac{1}{M-1} \left[2 \cdot \left[(\bm{z}_{k, \hat{m}, o} - \bar{\bm{z}}_{k, o}) \cdot \left({\frac{\partial \ \bm{z}_{k,\hat{m}, o}  }{ \partial \bm{z}_{k,\hat{m}, o}}}-{\frac{\partial  \bar{\bm{z}}_{k, o}  }{ \partial \bm{z}_{k,\hat{m}, o}}} \right)+ \textstyle\sum_{m=1}^M \mathbb{I}_{[m \neq \hat{m}]} (\bm{z}_{k, m, o} - \bar{\bm{z}}_{k, o}) \cdot \left({\frac{\partial  \bm{z}_{k,m, o}  }{ \partial \bm{z}_{k,\hat{m}, o}}}-{\frac{\partial  \bar{\bm{z}}_{k, o}  }{ \partial \bm{z}_{k,\hat{m}, o}}} \right) \right]\right] &=  \\ 
    \tfrac{-A}{2} \cdot \tfrac{1}{M-1} \left[2 \cdot \left[(\bm{z}_{k, \hat{m}, o} - \bar{\bm{z}}_{k, o}) \cdot \left(1-\tfrac{1}{M} \right)+ \textstyle\sum_{m=1}^M \mathbb{I}_{[m \neq \hat{m}]} (\bm{z}_{k, m, o} - \bar{\bm{z}}_{k, o}) \cdot \left(\tfrac{-1}{M} \right) \right]\right] &= \\ 
    \tfrac{-A}{M-1} \cdot \left[(\bm{z}_{k, \hat{m}, o} - \bar{\bm{z}}_{k, o}) \cdot \tfrac{M-1}{M} + \left( \left(\tfrac{\bm{z}_{k, \hat{m}, o}}{M} \right)- \bar{\bm{z}}_{k, o} + \tfrac{M-1}{M} \cdot \bar{\bm{z}}_{k, o} \right) \right]&= \\
    \tfrac{-A}{M-1} \cdot (\bm{z}_{k, \hat{m}, o} - \bar{\bm{z}}_{k, o})
    \end{aligned}
\end{equation}

\subsection{Proof for Eq.~\ref{eq:update:SGD}}~\label{sec:app:proof2}
\begin{equation*}
    \eta \cdot \nabla_{w_{\hat{m},o}} \ell_{div} = \eta \cdot {\frac{\partial \ell_{div} }{ \partial \bm{z}_{k,\hat{m},o}}}  \cdot {\frac{\partial  \bm{z}_{k,\hat{m},o} }{ \partial w_{\hat{m},o}}} = \eta \cdot \frac{-A}{M-1} \cdot (\bm{z}_{k, \hat{m},o} - \bar{\bm{z}}_{k,o}) \cdot b 
\end{equation*}


\end{document}